%% file: main.tex
\UseRawInputEncoding
\documentclass{article}
\usepackage{float}
\usepackage[margin=1.0in]{geometry}
\usepackage{abstract}
\usepackage{afterpage}
\usepackage[backend=bibtex,style=nature,natbib=true,maxbibnames=99,minalphanames=3]{biblatex}
\usepackage{hyperref}
\newcommand{\HandE}{H\texorpdfstring{\&}{&}E}
\usepackage{caption}
\usepackage{lmodern}
\usepackage{threeparttable}
\input{macros}

\usepackage{fontawesome5}
\usepackage{helvet}
\DeclareRobustCommand{\corrAuthor}{\textsuperscript{\faEnvelope[regular]}}
\renewcommand{\topfraction}{0.95}

\renewcommand{\floatpagefraction}{0.90}
\usepackage{soul}
\sethlcolor{yellow}
\usepackage{hyperref}
\usepackage{xcolor}
\hypersetup{breaklinks=true}

\hypersetup{
  colorlinks=true,
  urlcolor=blue,
  citecolor=ForestGreen,
  linkcolor=blue
}
\newcommand{\arialtitle}[1]{{\fontfamily{phv}\selectfont #1}}

\newif\ifpreprint
\preprintfalse    

\title{{\raggedright
\fontsize{15pt}{18pt}\selectfont \textbf{\arialtitle{CytoFormer: A Molecularly Supervised Cell Foundation Model for Histopathology Cell Classification
}}\par}}

\author{\name Jialu Yao$^{1,2,*}$, 
\name Songhao Li$^{1,2,*}$, 
\name Alina Yu$^{3}$, 
\name Zhi Huang$^{1,4,\corrAuthor}$\\\newline 
$^{1}$ Department of Pathology and Laboratory Medicine, University of Pennsylvania, Philadelphia, PA, USA\\
$^{2}$ Department of Electrical and Systems Engineering, University of Pennsylvania, Philadelphia, PA, USA\\
$^{3}$ Germantown Friends School, Philadelphia, PA, USA\\
$^{4}$ Department of Biostatistics,
Epidemiology \texorpdfstring{\&}{&} Informatics, University of Pennsylvania, Philadelphia, PA, USA\\
*~~~Equal contribution\\
\corrAuthor~Correspondence: Zhi Huang (\href{mailto:zhi.huang@pennmedicine.upenn.edu}{zhi.huang@pennmedicine.upenn.edu})
}

\begin{document}



\maketitle

\renewcommand{\abstractnamefont}{\normalfont\normalsize\bfseries}
\renewcommand{\abstracttextfont}{\normalfont\normalsize}

\renewcommand{\abstractname}{Abstract}
\begin{abstract}
\abstracttextfont
Identifying cell types directly from routine haematoxylin and eosin (\HandE{}) histology would enable single-cell analysis at scale, but training such models has relied on manual pathologist annotations, which are slow, expensive and unreliable for many cell types. We instead supervise morphology with molecules. Imaging-based spatial transcriptomics profiles individual cells \emph{in situ} on a section that can afterwards be stained with \HandE{}, so that molecular identity and morphology are observed for the same physical cell. We assembled $81$ such paired Xenium sections spanning $16$ organs, derived per-cell labels by clustering, marker-gene annotation, organ-wise human review and quality control, and mapped them onto the cell types commonly reported in each organ. This yielded $15.4$ million cells, each with a paired \HandE{} image patch and one of $23$ cell types, on which we trained CytoFormer, a cell foundation model with a multi-task, per-organ classification head. On spatially held-out tissue CytoFormer reached an accuracy of $0.85$ and a macro-F1 of $0.78$ across all $16$ organs, and its predictions reproduced the tissue architecture of an entire held-out section. The representation also transfers: with the encoder frozen, a linear head on CytoFormer features performed better than six pathology foundation models on four expert-annotated benchmarks, including on organs and cell types that were not part of pretraining. Finally, in an interactive active-learning setting, CytoFormer's embeddings are markedly more label-efficient than existing pathology foundation models, detecting normal epithelium amid look-alike tumour with an F1 of $0.82$ from only a few annotations and leading the strongest baseline by $0.13$ in F1. CytoFormer turns paired \HandE{} and spatial transcriptomics into a reusable, label-efficient representation for cell-level analysis of routine histology. Code and model checkpoints are available at \hyperlink{https://github.com/zhihuanglab/CytoFormer}{https://github.com/zhihuanglab/CytoFormer}.

\end{abstract}

\section{Introduction}

Haematoxylin and eosin (\HandE{}) staining is the basis of routine pathology diagnosis. Much of what a pathologist reads from an \HandE{} section is cellular: which cell types are present, in what proportion, and how they are arranged. Tumour cellularity, the density of tumour-infiltrating lymphocytes, the abundance of macrophages and plasma cells, and the state of the surrounding stroma and vasculature all carry diagnostic and prognostic weight. All of them can be read from the slide. In practice they are estimated by eye and reported semi-quantitatively, which is subjective and varies between observers. Typing every nucleus turns these judgements into counts, which are reproducible and can be compared across slides and cohorts.

Progress on this task has been limited by supervision rather than by architecture. In the past, public cell-classification resources were built from manual annotation by pathologists. Such annotation is slow and expensive, and for many cell types it is also unreliable: a macrophage, a fibroblast and a poorly differentiated tumour cell can be difficult to tell apart within a single \HandE{} nucleus. The largest expert-annotated datasets therefore contain on the order of $10^{5}$ nuclei and only a handful of coarse categories, as in PanNuke~\cite{gamper2019pannuke}, CoNSeP~\cite{graham2019hovernet} and MoNuSAC~\cite{verma2021monusac}, and their class definitions are not mutually consistent. Recently, pathology foundation models have advanced rapidly~\cite{Chen2024-sl,Lu2024-dz,Huang2023-di,Xiang2025-vd,Wang2024-hs,Xu2024-do,zimmermann2024virchow2}, and are widely used for downstream diagnostic tasks~\cite{yao2026melandx}, but they are pretrained and evaluated on patches and whole slides. Their supervision and representations can overlook cell-level features, so repurposing them for cell classification leaves the labelling bottleneck untouched.

Spatial transcriptomics offers a way around this bottleneck. Imaging-based platforms such as Xenium measure hundreds to thousands of transcripts \emph{in situ} at subcellular resolution and segment individual cells. The same section can then be stained with \HandE{} and imaged~\cite{janesick2023xenium}. Molecular identity and morphology are therefore available for the same physical cell, and the molecular channel can supervise the morphological one. Labels for \HandE{} can therefore be obtained without a pathologist drawing them, as long as the two images are registered and the molecular calls are checked. It also scales: a single Xenium section yields $10^{5}$ to $10^{6}$ typed cells, far more than a manually annotated cohort.

These labels are derived through a relatively comprehensive pipeline. Cell identity must be inferred from unsupervised clustering and marker interpretation, which is panel-dependent and error-prone. Segmentation errors and transcript spill-over generate doublets and ambient contamination~\cite{Marco-Salas2025-xz}. Artefact clusters are often tight and internally consistent, so they survive naive quality filters. Only a subset of cells can be confidently mapped onto a morphology after registration. A label set also has to be chosen, since the clusters found in one section do not by themselves define a taxonomy that holds across organs. Histology and spatial transcriptomics have been linked before, but existing models work at the level of spots or patches and predict or align gene expression~\cite{Chen2025-ee,Huang2025-bv}, rather than using the transcriptome as supervision for the identity of individual cells.

Here we present CytoFormer, a cell foundation model learned from paired \HandE{} and spatial transcriptomics. We curated $81$ Xenium sections spanning $16$ organs. Per-cell reference labels were derived by clustering, marker-based cluster annotation, cluster-level human quality control and a per-cell confidence gate, and were then mapped onto the cell types commonly reported in each organ, giving $23$ classes. Registering each section to its own \HandE{} image gave $15.4$ million cells with a paired $56\,\mu$m \HandE{} patch. On these we trained a cell encoder with a multi-task, per-organ classification head. CytoFormer classifies cell types on held-out sections across all $16$ organs, and its representation transfers. With the encoder frozen, a linear head on CytoFormer features performs better than six pathology foundation models on four expert-annotated benchmarks. The advantage is largest where supervision is scarce, and in an interactive active-learning setting CytoFormer detects normal epithelium amid look-alike tumour from only a few hundred annotations.


\renewcommand{\floatpagefraction}{0.55}
\renewcommand{\topfraction}{0.85}

\section{Results}
\label{sec:benchmarks}



\begin{figure}[p]
    \centering
    \includegraphics[width=0.95\textwidth]{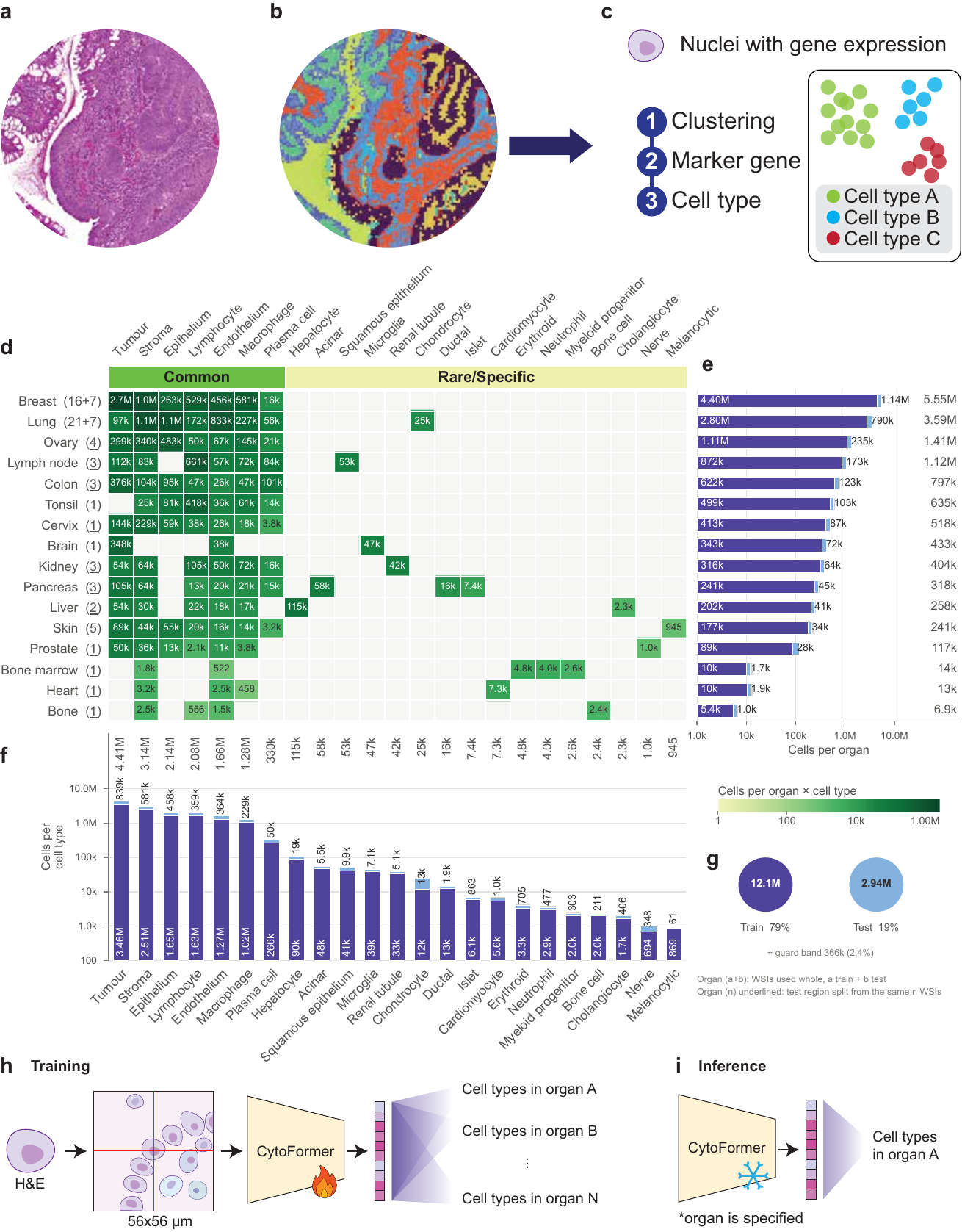}
\end{figure}

\begin{figure}[!t]
        \captionsetup{
          format=plain,
          justification=raggedright,
          singlelinecheck=false,
          margin=0pt,
          width=\dimexpr\textwidth\relax
        }
    \caption[\textbf{A pan-tissue single-cell \HandE{} dataset built from paired spatial transcriptomics.}]%
    {\textbf{A pan-tissue single-cell \HandE{} dataset built from paired spatial transcriptomics.}
    \textbf{(a)} \HandE{} image of a tissue section that was also profiled with Xenium.
    \textbf{(b)} Xenium cell map of the same section.
    \textbf{(c)} Cell types are derived by clustering, marker-gene annotation and human review, and transferred to the paired \HandE{} morphology.
    \textbf{(d)} Cell types recovered in each organ, coloured and labelled by the number of cells.
    \textbf{(e)} Cells per organ, split into training (dark) and test (light) cells.
    \textbf{(f)} Cells per cell type, split into training and test cells; both axes are logarithmic.
    \textbf{(g)} Overall training and test split.
    \textbf{(h)} During training, a $56\times56\,\mu$m crop around each nucleus is encoded by CytoFormer and classified by a per-organ head.
    \textbf{(i)} At inference the encoder is frozen and the prediction is made by the head of the specified organ.}
    \label{fig:figure1}
\end{figure}

\subsection{Curate a pan-tissue single-cell \HandE{} dataset from paired spatial transcriptomics}
\label{sec:dataset}

To train a cell-level model without manual annotation, we built a single-cell \HandE{} dataset in which every cell is typed by its own transcriptome. Imaging-based spatial transcriptomics profiles single cells \emph{in situ}, and the same section can afterwards be stained with \HandE{} and imaged. Each physical cell is therefore observed twice: as morphology on the \HandE{} image (\textbf{Figure~\ref{fig:figure1}a}), and as an expression profile in the Xenium cell map (\textbf{Figure~\ref{fig:figure1}b}). We assembled public human Xenium experiments that carry a paired \HandE{} image of the same section, drawn from the 10x Genomics data portal and from HEST-1k~\cite{jaume2024hest}. We then assigned a cell type to every cell in three steps: unsupervised clustering of the expression profiles, annotation of each cluster from its marker genes, and organ-wise human review of every cluster (\textbf{Figure~\ref{fig:figure1}c}). Each type was then transferred to the matched \HandE{} morphology. After quality control, per-cell confidence filtering and registration, the dataset contains $15{,}422{,}352$ cells from $81$ sections and $16$ organs. Every cell carries a $56\,\mu$m \HandE{} patch, a cell type and an organ label. Of the sections, $54$ come from the 10x Genomics portal and contribute $12{,}078{,}610$ cells ($78.3$\%); the other $27$ come from HEST-1k and contribute $3{,}343{,}742$ cells ($21.7$\%).

The cell types split into common classes and organ-specific classes (\textbf{Figure~\ref{fig:figure1}d}). Seven common classes are shared by most organs: Tumour, Stroma, Epithelium, Lymphocyte, Endothelium, Macrophage and Plasma cell. Together they cover $97.5$\% of the dataset. Endothelium is the only class found in all $16$ organs. Stroma appears in $15$ organs, and Lymphocyte and Macrophage in $13$ each. The other $16$ classes are rare, and each appears in a single tissue: Hepatocyte and Cholangiocyte in liver, Acinar, Ductal and Islet in pancreas, Renal tubule in kidney, Microglia in brain, Squamous epithelium in lymph node, Chondrocyte in lung, Cardiomyocyte in heart, Erythroid, Neutrophil and Myeloid progenitor in bone marrow, Bone cell in bone, Nerve in prostate, and melanocytic in skin. The cell types were chosen per organ rather than globally: for each tissue we used the types that are commonly reported in it. Each organ therefore has its own list, with a median of seven cell types and up to nine in pancreas.

The dataset covers $16$ organs: breast, lung, ovary, lymph node, colon, tonsil, cervix, brain, kidney, pancreas, liver, skin, prostate, bone marrow, heart and bone (\textbf{Figure~\ref{fig:figure1}e}). Breast is the largest, with $5{,}546{,}376$ cells ($36.0$\%) from $23$ sections. Lung is the second largest, with $3{,}592{,}765$ cells ($23.3$\%) from $28$ sections. Together they hold $59.3$\% of all cells, and they are the only two organs with many comparable sections. The other $14$ organs are each represented by one to five sections, and range from $1{,}405{,}609$ cells in ovary to $6{,}928$ cells in bone.

The cell counts are long-tailed and span more than three orders of magnitude (\textbf{Figure~\ref{fig:figure1}f}). Tumour is the largest class, with $4{,}408{,}081$ cells ($28.6$\%). Stroma is the second largest, with $3{,}139{,}069$ cells ($20.4$\%). The two together make up almost half of the dataset. The remaining common classes hold between $329{,}965$ cells (Plasma cell) and $2{,}141{,}342$ cells (Epithelium). The organ-specific classes are much smaller, from $114{,}733$ Hepatocytes down to $1{,}042$ Nerve cells and $945$ melanocytic cells. This imbalance comes from sampling whole tissue sections. We handle it during training rather than by resampling the dataset.

Neighbouring cells share tissue context, so we split training and test cells by space rather than at random (\textbf{Figure~\ref{fig:figure1}g}). For breast and lung we held out whole sections, $7$ of $23$ and $7$ of $28$. The other $14$ organs have only one to five sections each. For these we cut every section into spatial blocks and held out dispersed blocks. We then discarded a $112\,\mu$m guard band along each train--test boundary, so that no test cell shares a field of view with a training cell. This gives $12{,}111{,}769$ training cells ($79$\%), $2{,}944{,}356$ test cells ($19$\%) and $366{,}227$ guard-band cells ($2.4$\%). Every cell type is present in both splits.

\subsection{Train a cell foundation model across 16 organs}
\label{sec:model}

CytoFormer classifies one cell at a time. For every nucleus we crop a $56\times56\,\mu$m field of view centred on its centroid and resize it to $224\times224$ pixels (\textbf{Figure~\ref{fig:figure1}h}). The crop keeps the nucleus at diagnostic resolution and still shows its immediate neighbours, which is the trade-off examined in \textbf{Section~\ref{sec:ablation}}. The image encoder is a ViT-giant transformer that outputs a $1536$-dimensional embedding per cell. It is initialised from UNI2-h~\cite{Chen2024-sl} and fine-tuned end-to-end, and its output is layer-normalised before classification.

A cell type does not look the same in every organ. A macrophage in lung alveoli, a Kupffer cell in liver sinusoids and a tumour-associated macrophage in breast stroma share a lineage but differ in shape and in the tissue around them. Many cell types are also organ-specific, such as hepatocytes in liver or islet cells in pancreas. One classifier shared by all organs would have to cover every appearance at once and rule out the types that cannot occur.

We therefore learn one linear classifier per organ on top of a single shared encoder (\textbf{Figure~\ref{fig:figure1}h}). Head sizes range from three to nine classes, and the organ identifier selects which head is applied. In a clinical setting the organ is usually known, so it is available at inference (\textbf{Figure~\ref{fig:figure1}i}). The heads are organ-specific, but the encoder is not. It is trained jointly on all $16$ organs, so the $16$ classifiers are not independent models but share one representation. Features that recur in every tissue, such as the small round nucleus of a lymphocyte or the flat profile of a vessel lining, are therefore learned from all organs together, and each head only has to separate the cell types of its own organ.

The encoder and all $16$ heads were trained jointly on the $12{,}111{,}769$ training cells, and we kept the checkpoint with the best macro-F1 on the held-out split. Training settings and hyperparameters are described in \textbf{Methods}.

\begin{figure}[!tp]
    \centering
    \includegraphics[width=0.95\textwidth]{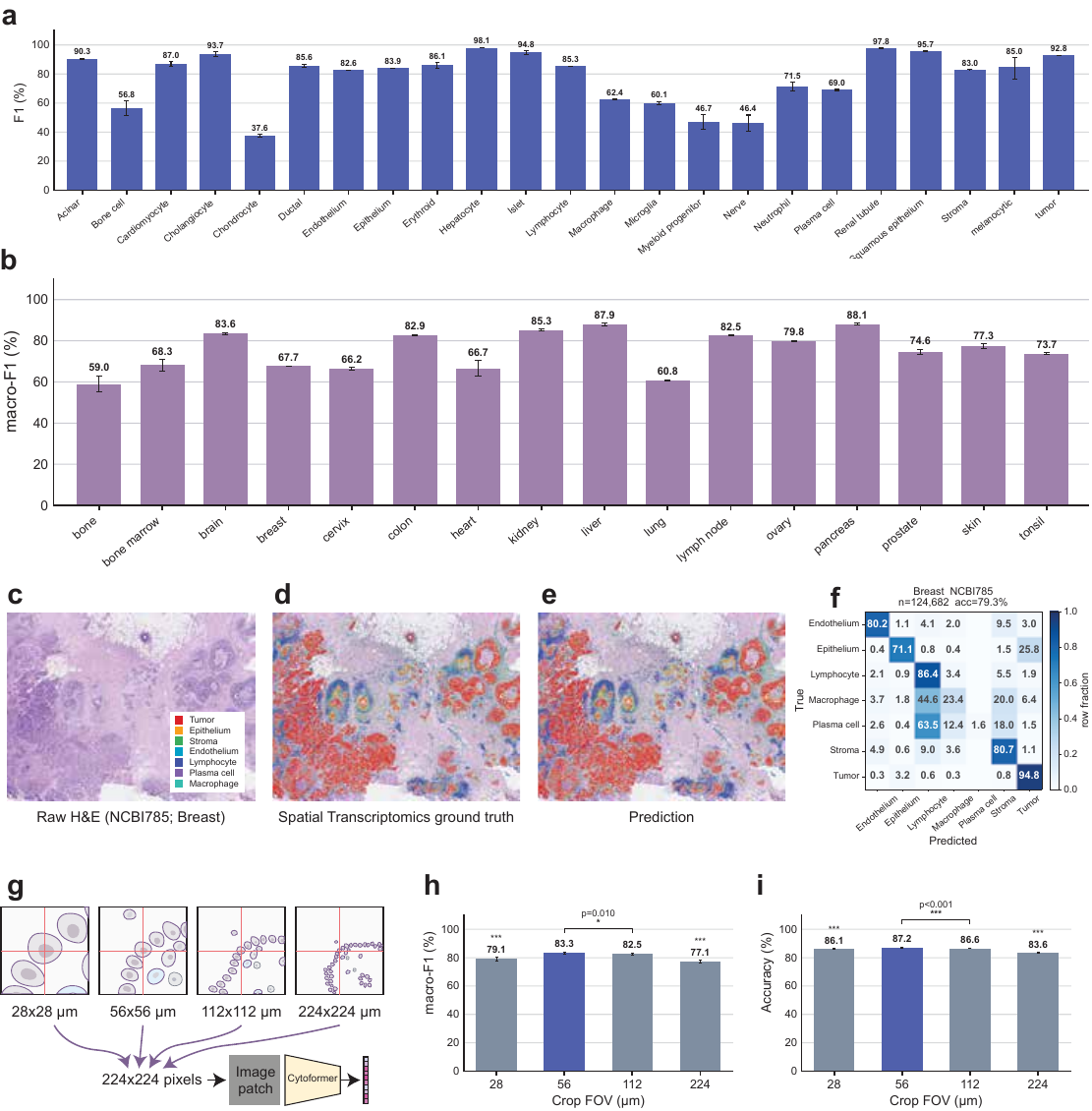}
        \caption[\textbf{CytoFormer classifies cell types on held-out \HandE{} tissue across 16 organs.}]%
    {\textbf{CytoFormer classifies cell types on held-out \HandE{} tissue across 16 organs.}
    \textbf{(a)} F1 per cell type and \textbf{(b)} macro-F1 per organ on the spatially held-out test split.
    \textbf{(c--e)} A breast section held out in its entirety: \textbf{(c)} raw \HandE{}, \textbf{(d)} the cell-type reference derived from the paired spatial transcriptomics and \textbf{(e)} the CytoFormer prediction.
    \textbf{(f)} Confusion matrix for that section.
    \textbf{(g)} Crop field-of-view ablation: one fixed set of cells is cropped at each field of view and resized to $224\times224$ pixels.
    \textbf{(h)} Macro-F1 and \textbf{(i)} accuracy at each crop field of view.
    Error bars represent $95$\% confidence intervals from bootstrap resampling over test cells. Statistical significance against the $56\,\mu$m setting was assessed using a paired two-sided bootstrap test: $^{*}p<0.05$, $^{**}p<0.01$, $^{***}p<0.001$. A bracket is drawn only between the $56\,\mu$m setting and the second-best setting and gives their exact $p$-value; the other settings are marked with asterisks only.}
    \label{fig:figure2}
\end{figure}

\subsection{CytoFormer recovers cell types and tissue architecture on held-out sections}
\label{sec:indistribution}

We first evaluated CytoFormer on the spatially held-out split, which contains $2{,}944{,}356$ cells from all $16$ organs. The model reached an accuracy of $0.846$ and a macro-F1 of $0.779$. Per-class F1 varied widely (\textbf{Figure~\ref{fig:figure2}a}). Cell types with a distinctive morphology were recognised almost perfectly: Hepatocyte reached $98.1$, Renal tubule $97.8$, Squamous epithelium $95.7$, Islet $94.8$, Cholangiocyte $93.7$, tumour $92.8$ and Acinar $90.3$. The five most abundant classes were also recovered reliably, with tumour at $92.8$, Lymphocyte at $85.3$, Epithelium at $83.9$, Endothelium at $82.6$ and Stroma at $83.0$. The weakest classes were Chondrocyte ($37.6$), Nerve ($46.4$), Myeloid progenitor ($46.7$), Bone cell ($56.8$), Microglia ($60.1$) and Macrophage ($62.4$).

Accuracy was driven by morphological distinctiveness rather than by how many cells a class contributes. Several very rare classes were classified well, including melanocytic at $85.0$ from only $61$ test cells, Cholangiocyte at $93.7$ from $406$, Erythroid at $86.1$ from $705$ and Islet at $94.8$ from $863$. Conversely, two of the weakest classes are not rare at all: Macrophage has $228{,}718$ test cells and reaches $62.4$, and Chondrocyte has $12{,}508$ and reaches $37.6$. The classes that fail are those whose nuclei resemble another class in the same tissue, in particular the mononuclear cells, where macrophages and plasma cells are drawn towards lymphocytes.

Performance by organ followed the same pattern (\textbf{Figure~\ref{fig:figure2}b}). Macro-F1 was highest in organs built from a few visually distinct compartments, reaching $88.1$ in pancreas, $87.9$ in liver, $85.3$ in kidney, $83.6$ in brain and $82.9$ in colon. It was lowest in the organs with the fewest cells or the most confusable stromal and immune mixtures, at $59.0$ in bone, $60.8$ in lung, $66.2$ in cervix and $66.7$ in heart. Accuracy was considerably higher than macro-F1 wherever one class dominates the organ, for example $86.0$ against $67.7$ in breast, $91.5$ against $74.6$ in prostate and $77.0$ against $60.8$ in lung. Per-organ and per-class accuracy, precision and recall are reported in \textbf{Extended Data Figure~\ref{fig:extfigure1}}.

To show what the predictions look like in practice, we applied the model to a complete breast section that was held out from training. \textbf{Figure~\ref{fig:figure2}c} shows the raw \HandE{}, \textbf{Figure~\ref{fig:figure2}d} the spatial-transcriptomic reference and \textbf{Figure~\ref{fig:figure2}e} the CytoFormer prediction for the same region. All $124{,}682$ cells of the section were classified, with an accuracy of $79.3$\%. Comparing the two maps in \textbf{Figure~\ref{fig:figure2}d,e} shows that the predicted tissue architecture matches the reference. The same ducts, the same stromal bands and the same immune aggregates appear in the same places, and the visible disagreements are at the level of individual cells rather than of structures. \textbf{Figure~\ref{fig:figure2}f} gives the confusion matrix for this same section, and shows where the model is reliable and where it is not. Tumour, Lymphocyte, Stroma and Endothelium were recovered at $94.8$\%, $86.4$\%, $80.7$\% and $80.2$\% of their true cells. Most of the error mass sits in the mononuclear compartment, with $44.6$\% of macrophages and $63.5$\% of plasma cells predicted as lymphocytes, and a further $25.8$\% of epithelial cells predicted as tumour. These are the distinctions that motivated merging cell types in the first place, and they remain the limiting factor rather than a diffuse loss of accuracy.

\subsection{Ablation study on the crop field of view}
\label{sec:ablation}

Cell classification on \HandE{} requires both the nucleus itself and enough of its neighbourhood to place it in context, and at a fixed model input size these two demands are in direct conflict: a larger crop adds context but lowers the effective resolution. We tested this trade-off directly by fixing one set of cells ($500$k training and $100$k test cells stratified over $10$ organs and $17$ cell types) and cropping the same cells at $28$, $56$, $112$ and $224\,\mu$m, always resized to $224$ pixels, so that the crop window is the only variable (\textbf{Figure~\ref{fig:figure2}g}).

A $56\,\mu$m crop was the best setting (macro-F1 $79.1$, $83.3$, $82.5$ and $77.1$ and accuracy $86.1$, $87.2$, $86.6$ and $83.6$ for $28$, $56$, $112$ and $224\,\mu$m; \textbf{Figure~\ref{fig:figure2}h,i}). Both metrics dropped at the two extremes. Under a paired bootstrap over the $100$k shared test cells, $56\,\mu$m was better than $28$ and $224\,\mu$m ($p<0.001$) and better than $112\,\mu$m ($p=0.010$). A $28\,\mu$m crop is close to the nucleus alone and loses the tissue context that separates, for example, stroma from endothelium; a $224\,\mu$m crop contains a whole neighbourhood but at $1\,\mu$m per pixel the target nucleus is no longer resolved. A $56\,\mu$m window, corresponding to $0.25\,\mu$m per pixel at $224$ pixels, keeps the nucleus at diagnostic resolution while still including its immediate neighbours, and we adopt it throughout.

\begin{figure}[p]
    \centering
    \includegraphics[width=0.95\textwidth]{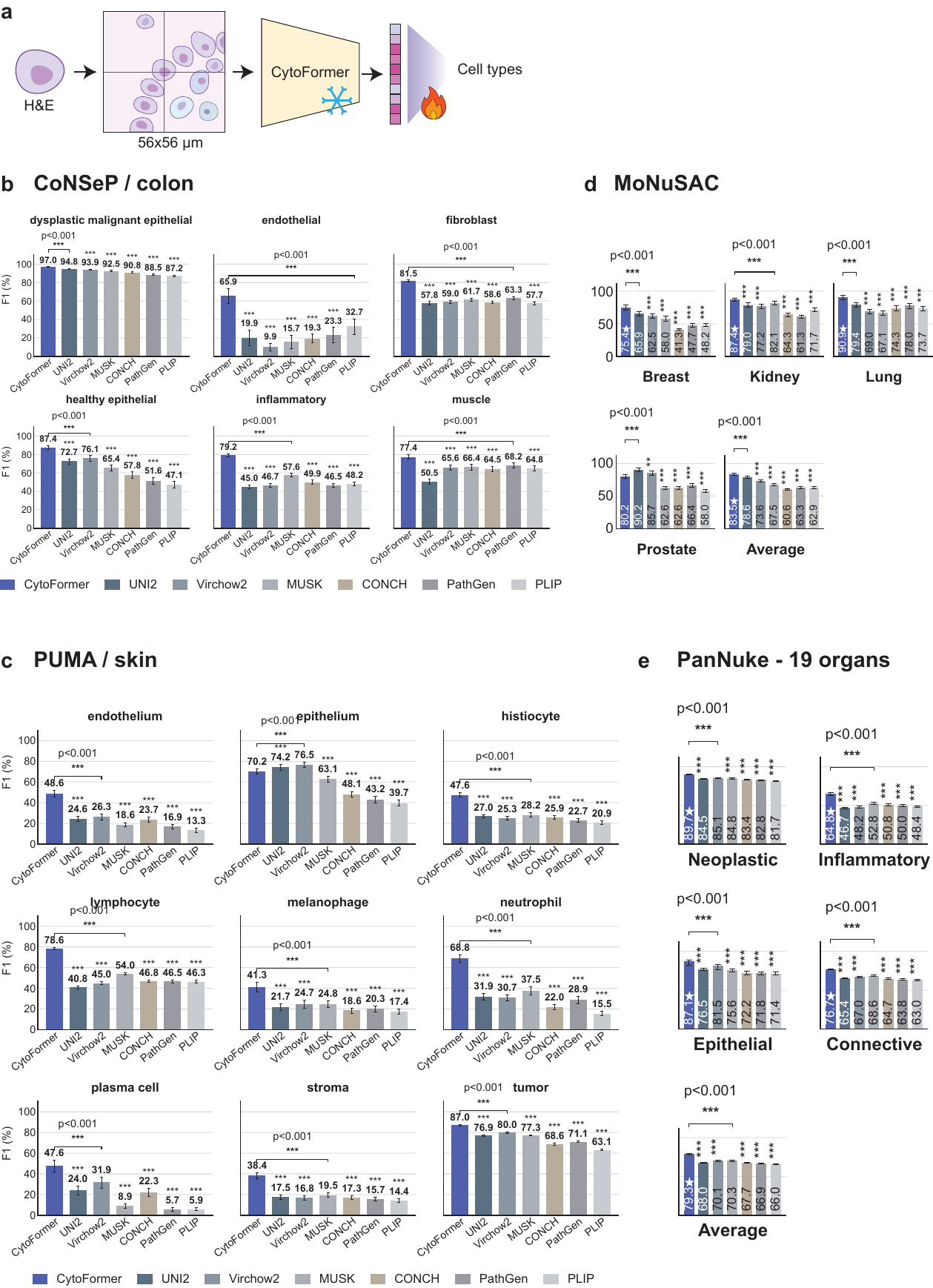}
\end{figure}

\begin{figure}[!t]
        \caption[\textbf{Linear fine-tuning of CytoFormer and six pathology foundation models on four public benchmarks.}]%
    {\textbf{Linear fine-tuning of CytoFormer and six pathology foundation models on four public benchmarks.}
    \textbf{(a)} Linear fine-tuning setup. A $56\,\mu$m crop around each nucleus is encoded by the frozen backbone (snowflake) and classified by a linear head, which is the only trainable part (flame).
    \textbf{(b)} F1 per cell type on CoNSeP (colon).
    \textbf{(c)} F1 per cell type on PUMA (melanoma skin).
    \textbf{(d)} Macro-F1 per organ on MoNuSAC and their average.
    \textbf{(e)} F1 per cell type on PanNuke and the overall average, each averaged over the $19$ organs.
    Error bars represent $95$\% confidence intervals from bootstrap resampling over test cells. Statistical significance was assessed using a paired two-sided bootstrap test: $^{*}p<0.05$, $^{**}p<0.01$, $^{***}p<0.001$. A bracket is drawn only between CytoFormer and the second-best method and gives their exact $p$-value; the remaining methods are marked with asterisks only.}
    \label{fig:figure3}
\end{figure}

\subsection{Linear fine-tuning on public cell-classification benchmarks}
\label{sec:linprobe}

We next tested how well CytoFormer performs under linear fine-tuning. The encoder was frozen and only a linear head on top of it was trained (\textbf{Figure~\ref{fig:figure3}a}). We compared it with six pathology foundation models used in the same way: UNI2-h~\cite{Chen2024-sl}, Virchow2~\cite{zimmermann2024virchow2}, MUSK~\cite{Xiang2025-vd}, CONCH~\cite{Lu2024-dz}, PathGen~\cite{sun2025pathgen} and PLIP~\cite{Huang2023-di}. The evaluation used four datasets annotated by pathologists and independent of our training data: PanNuke~\cite{gamper2019pannuke} ($186{,}836$ cells, $19$ organs, $4$ classes), CoNSeP~\cite{graham2019hovernet} ($23{,}400$ cells, colon, $6$ classes), MoNuSAC~\cite{verma2021monusac} ($46{,}901$ cells, $4$ organs, $4$ classes) and PUMA~\cite{schuiveling2025puma} ($95{,}389$ cells, melanoma skin, $9$ classes). All backbones were evaluated on the same cells, with the same $56\,\mu$m crops and the same class-balanced linear head.

Each dataset was evaluated organ by organ, giving $25$ settings in total: $19$ organs in PanNuke, four in MoNuSAC, and one each in CoNSeP and PUMA. CytoFormer gave the best representation in $24$ of these $25$ settings, and led on every dataset. On CoNSeP, which has six classes in colon, CytoFormer had the highest F1 on all six (\textbf{Figure~\ref{fig:figure3}b}). Against the second-best method of each class, it reached $97.0$ against $94.8$ for UNI2-h on dysplastic malignant epithelium, $87.4$ against $76.1$ for Virchow2 on healthy epithelium, $81.5$ against $63.3$ for PathGen on fibroblasts, $79.2$ against $57.6$ for MUSK on inflammatory cells, $77.4$ against $68.2$ for PathGen on muscle and $65.9$ against $32.7$ for PLIP on endothelial cells, in every case with $p<0.001$. At the organ level it reached a macro-F1 of $81.4$ against $59.9$ for MUSK ($p<0.001$). On PUMA, which has nine classes in melanoma skin, CytoFormer was highest on eight of them (\textbf{Figure~\ref{fig:figure3}c}). Against the second-best method of each class, it reached $87.0$ against $80.0$ for Virchow2 on tumour, $78.6$ against $54.0$ for MUSK on lymphocytes, $68.8$ against $37.5$ for MUSK on neutrophils, $48.6$ against $26.3$ for Virchow2 on endothelium, $47.6$ against $28.2$ for MUSK on histiocytes, $47.6$ against $31.9$ for Virchow2 on plasma cells, $41.3$ against $24.8$ for MUSK on melanophages and $38.4$ against $19.5$ for MUSK on stroma, again all with $p<0.001$. The one class on which CytoFormer did not lead is epithelium, where it still ranked among the top methods. At the organ level CytoFormer reached $58.7$ against $39.7$ for Virchow2 ($p<0.001$).

On MoNuSAC, CytoFormer led in three of the four organs, with $75.4$ against $65.9$ for UNI2-h in breast, $87.4$ against $82.1$ for MUSK in kidney and $90.9$ against $79.4$ for UNI2-h in lung, all with $p<0.001$ (\textbf{Figure~\ref{fig:figure3}d}). The one organ in which CytoFormer did not lead is prostate, where it still ranked among the top methods. Averaged over the four organs it reached $83.5$ against $78.6$ for UNI2-h ($p<0.001$). On PanNuke, averaged over its $19$ organs, it was highest on all four classes, with $89.7$ against $85.1$ for Virchow2 on neoplastic cells, $87.1$ against $81.5$ for Virchow2 on epithelial cells, $76.7$ against $68.6$ for MUSK on connective cells and $64.8$ against $52.8$ for MUSK on inflammatory cells, and averaged $79.3$ against $70.3$ for MUSK, all with $p<0.001$ (\textbf{Figure~\ref{fig:figure3}e}). Across all $25$ settings CytoFormer averaged $79.2$ macro-F1, against $71.4$ for the second-best method of each setting.

CytoFormer therefore performed well on all four datasets, at both the class and the organ level. The advantage was not restricted to the organs and cell types it had been trained on, and CytoFormer performed just as well on those it had never seen. Nine of PanNuke's $19$ organs are not among the organs seen during pretraining, among them bladder, oesophagus, stomach, thyroid, testis and adrenal gland, and CytoFormer led on every one of them, by the same margin as on the organs it had seen. It also led on cell types that its own label space never contained as separate classes, such as melanophages in PUMA. This indicates that what the model has learned is generic cell-level information, the appearance of the nucleus together with the tissue immediately around it, rather than a fixed set of organ-specific categories.

\begin{figure}[p]
    \centering
    \includegraphics[width=0.95\textwidth]{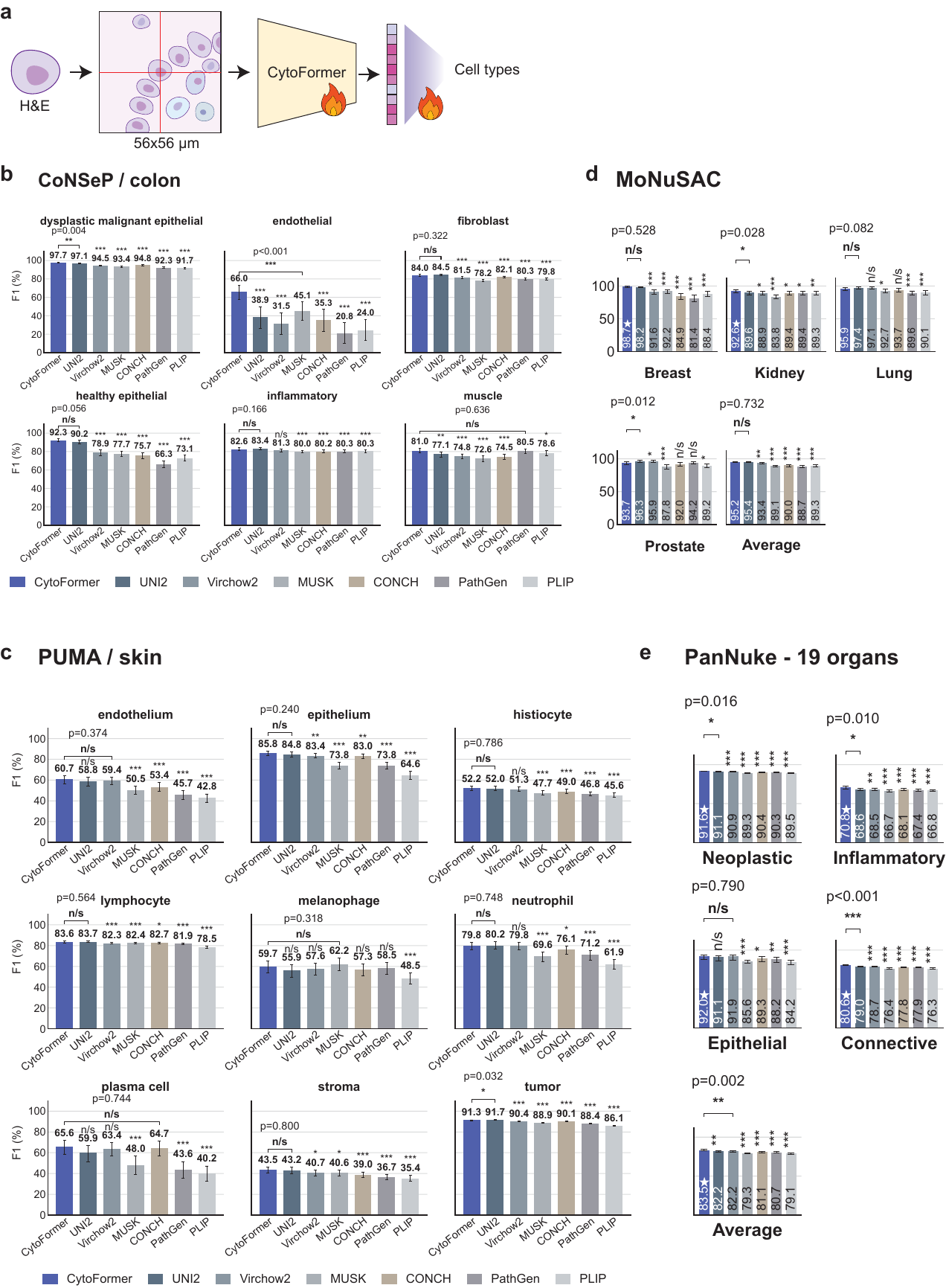}
\end{figure}

\begin{figure}[!t]
        \caption[\textbf{Full fine-tuning of CytoFormer and six pathology foundation models on four public benchmarks.}]%
    {\textbf{Full fine-tuning of CytoFormer and six pathology foundation models on four public benchmarks.}
    \textbf{(a)} Full fine-tuning setup. A $56\,\mu$m crop around each nucleus is encoded by the backbone and classified by a head, and both are trained end-to-end (flames).
    \textbf{(b)} F1 per cell type on CoNSeP (colon).
    \textbf{(c)} F1 per cell type on PUMA (melanoma skin).
    \textbf{(d)} Macro-F1 per organ on MoNuSAC and their average.
    \textbf{(e)} F1 per cell type on PanNuke and the overall average, each averaged over the $19$ organs.
    Error bars represent $95$\% confidence intervals from bootstrap resampling over test cells. Statistical significance was assessed using a paired two-sided bootstrap test: $^{*}p<0.05$, $^{**}p<0.01$, $^{***}p<0.001$. A bracket is drawn only between CytoFormer and the second-best method and gives their exact $p$-value; the remaining methods are marked with asterisks only.}
    \label{fig:figure4}
\end{figure}

\subsection{Full fine-tuning on public cell-classification benchmarks}
\label{sec:finetune}

We then repeated the comparison with every backbone fine-tuned end-to-end (\textbf{Figure~\ref{fig:figure4}a}). This evaluation asks a practical question: when choosing a pretrained model for \HandE{} cell classification, which backbone provides the strongest starting point and can achieve the best performance after full fine-tuning? The datasets, the cells, the crops and the training recipe were the same for all methods, so the only difference remains the backbone the model starts from.

We fully fine-tuned every model on each benchmark. CytoFormer continued to perform very well. On CoNSeP it had the highest F1 on four of the six classes (\textbf{Figure~\ref{fig:figure4}b}), with $97.7$ against $97.1$ for UNI2-h on dysplastic malignant epithelium ($p=0.004$), $92.3$ against $90.2$ for UNI2-h on healthy epithelium ($p=0.056$), $81.0$ against $80.5$ for PathGen on muscle ($p=0.636$) and $66.0$ against $45.1$ for MUSK on endothelial cells ($p<0.001$); on fibroblasts and inflammatory cells CytoFormer was not the highest, but in both cases the difference from the best method was not significant ($p=0.322$ and $p=0.166$). At the organ level CytoFormer reached a macro-F1 of $83.9$ against $78.5$ for UNI2-h ($p<0.001$). On PUMA it had the highest F1 on five of the nine classes (\textbf{Figure~\ref{fig:figure4}c}), among them epithelium at $85.8$ against $84.8$ and plasma cells at $65.6$ against $64.7$. On the remaining four classes CytoFormer was not the highest; on three of them the difference from the best method was not significant, and on tumour it stayed among the top methods. At the organ level CytoFormer reached $69.1$ against $67.8$ for UNI2-h ($p=0.008$).

On MoNuSAC, CytoFormer had the highest macro-F1 in breast ($98.7$ against $98.2$ for UNI2-h, $p=0.528$) and kidney ($92.6$ against $89.6$ for UNI2-h, $p=0.028$). In lung it was not the highest, but the difference from the best method was not significant ($p=0.082$), and in prostate it stayed among the top methods. Averaged over the four organs CytoFormer reached $95.2$, with no significant difference from the best method ($p=0.732$; \textbf{Figure~\ref{fig:figure4}d}). On PanNuke, averaged over its $19$ organs, CytoFormer was highest on all four classes, with $91.6$ against $91.1$ for UNI2-h on neoplastic cells ($p=0.016$), $92.0$ against $91.9$ for Virchow2 on epithelial cells ($p=0.790$), $80.6$ against $79.0$ for UNI2-h on connective cells ($p<0.001$) and $70.8$ against $68.6$ for UNI2-h on inflammatory cells ($p=0.010$), and averaged $83.5$ against $82.2$ for Virchow2 ($p=0.002$; \textbf{Figure~\ref{fig:figure4}e}).

CytoFormer therefore still performed well under full fine-tuning on every dataset. It stayed clearly ahead on the difficult classes, such as endothelial cells in CoNSeP and connective cells in PanNuke, even after every backbone had been trained on the target labels. This confirms that pretraining on our dataset is effective, and that CytoFormer is a strong starting point for cell classification on \HandE{}.

\begin{figure}[p]
    \centering
    \includegraphics[width=0.95\textwidth]{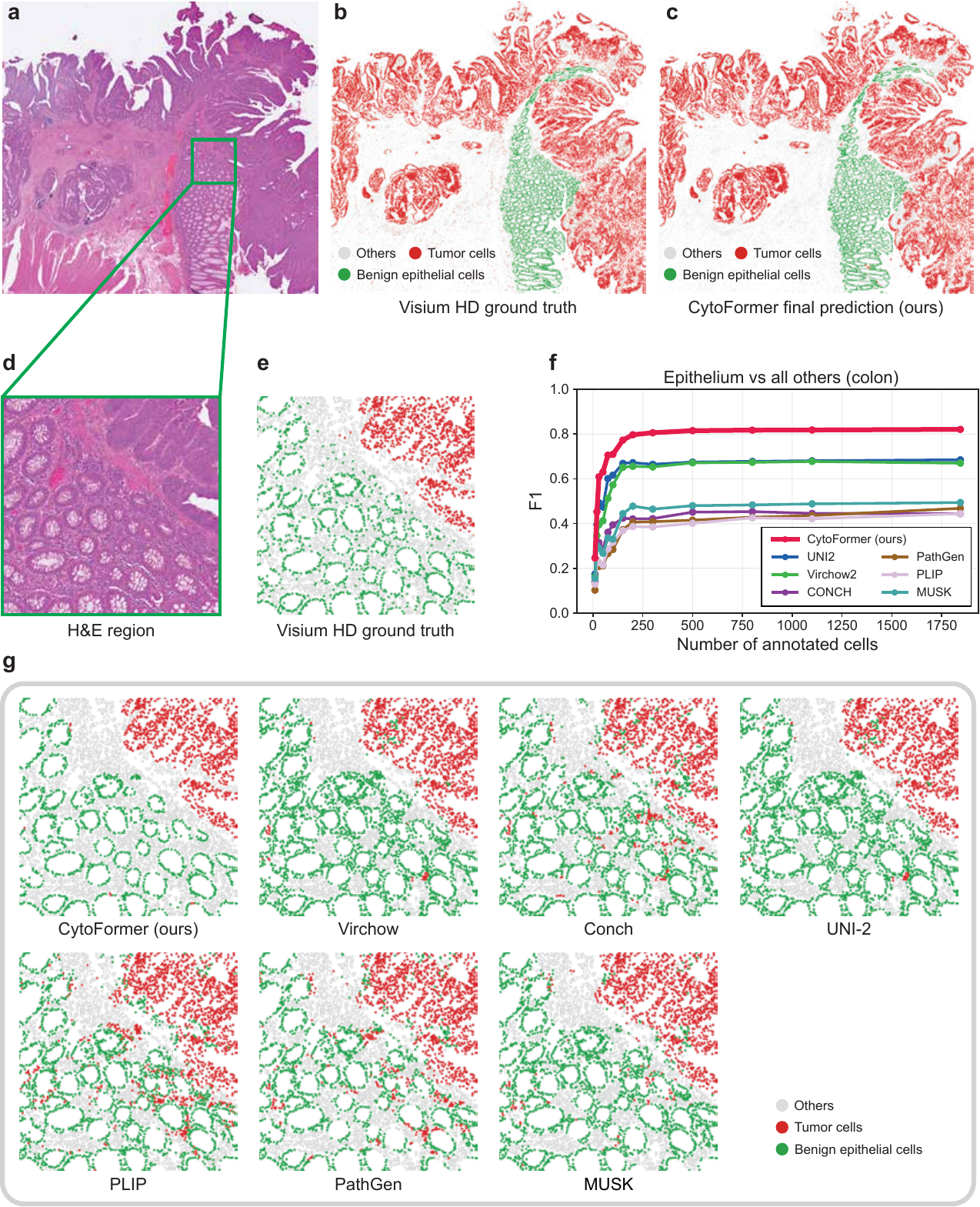}
\end{figure}

\begin{figure}[!t]
        \captionsetup{
          format=plain,
          justification=raggedright,
          singlelinecheck=false,
          margin=0pt,
          width=\dimexpr\textwidth\relax
        }
    \caption[\textbf{CytoFormer embeddings enable label-efficient active learning for cell classification on VisiumHD.}]%
    {\textbf{CytoFormer embeddings enable label-efficient active learning for cell classification on VisiumHD.}
    For every foundation model an identical logistic-regression head is trained on the same user annotations and evaluated on the same held-out cells, so that the embedding is the only variable.
    \textbf{(a)} \HandE{} whole-slide image of the section.
    \textbf{(b)} VisiumHD ground-truth overlay, with Negative control shown in grey, Tumor in red and Epithelium in green.
    \textbf{(c)} Active-learning prediction from CytoFormer (ours), with the Epithelium
    prediction confined to the normal crypt region.
    \textbf{(d)} \HandE{} of a zoomed region of interest (ROI) spanning the benign--malignant boundary.
    \textbf{(e)} Ground truth of the zoomed region of interest spanning the benign--malignant boundary.
    \textbf{(f)} Active-learning learning curves (F1 for detecting normal Epithelium versus all
    other cells) as a function of the number of annotated cells, averaged over five random subsets;
    CytoFormer (bold) is highest throughout and improves most with additional annotations.
    \textbf{(g)} Per-model prediction overlays within the ROI: CytoFormer places Epithelium precisely on the normal crypts, whereas frozen foundation models over-call epithelium across malignant
    and non-epithelial cells.}
    \label{fig:figure5}
\end{figure}

\subsection{CytoFormer provides more label-efficient representations for interactive cell annotation}
\label{activelearning}
A practical foundation model should let a pathologist obtain accurate cell typing from only a handful of annotations. We therefore benchmarked CytoFormer against existing pathology foundation models as the representation for interactive, active-learning cell annotation on our TissueLab platform~\cite{Li2025-kj}, focusing on label efficiency.

To evaluate it, we used a colorectal cancer VisiumHD section (\textbf{Figure~\ref{fig:figure5}a}) and framed the task as detecting normal epithelium, which is hard because benign crypt epithelium and adenocarcinoma look alike on H\&E and share most marker genes. We derived per-cell ground truth from the paired VisiumHD transcriptome together with the crypt structure (\textbf{Figure~\ref{fig:figure5}b}). For every foundation model baseline we trained an identical logistic-regression head on the same user annotations and evaluated normal-epithelium detection, Epithelium versus all other cells, across the section ($10{,}627$ positive and $151{,}075$ negative ground-truth cells), so that the embedding was the only variable.

CytoFormer detected normal epithelium most accurately, reaching an F1 of $0.82$ and confining the Epithelium prediction to the normal crypt region (\textbf{Figure~\ref{fig:figure5}c}). All other foundation-model
baselines scored markedly lower, with UNI2-h the strongest at F1 $0.69$, followed by Virchow2 at $0.67$, MUSK at $0.49$, PathGen at $0.47$, and CONCH and PLIP at $0.44$, so that CytoFormer led the best baseline by about $0.13$ in F1. Zooming into a region of interest at the benign--malignant boundary (\textbf{Figure~\ref{fig:figure5}d}, with ground truth in \textbf{Figure~\ref{fig:figure5}e}) made this difference concrete. The advantage was largest in the low-annotation regime that active learning operates in. Across the annotation schedule, the CytoFormer learning curve rose fastest and highest (\textbf{Figure~\ref{fig:figure5}f} and \textbf{Extended Data Figure~\ref{fig:extfigure2}}), reaching most of its final F1 within the first ${\sim}200$ annotations, whereas the baselines plateaued well below it and did not close the gap even at the largest annotation budget. Per-model overlays within this region (\textbf{Figure~\ref{fig:figure5}g}) confirmed that CytoFormer placed Epithelium precisely on the normal crypts while the baselines scattered it across malignant and non-epithelial cells. Because every model used an identical classifier and differed only in its embedding, this advantage is attributable to the representation, showing that CytoFormer yields linearly separable, label-efficient cell features that make detecting subtle cell states such as normal epithelium amid look-alike tumour practical from only a few annotations.


\setcounter{figure}{0}
\renewcommand{\figurename}{Extended Data Figure}
\renewcommand{\floatpagefraction}{0.45}

\begin{figure}[p]
    \centering
    \includegraphics[width=0.95\textwidth,trim=0 0 0 40,clip]{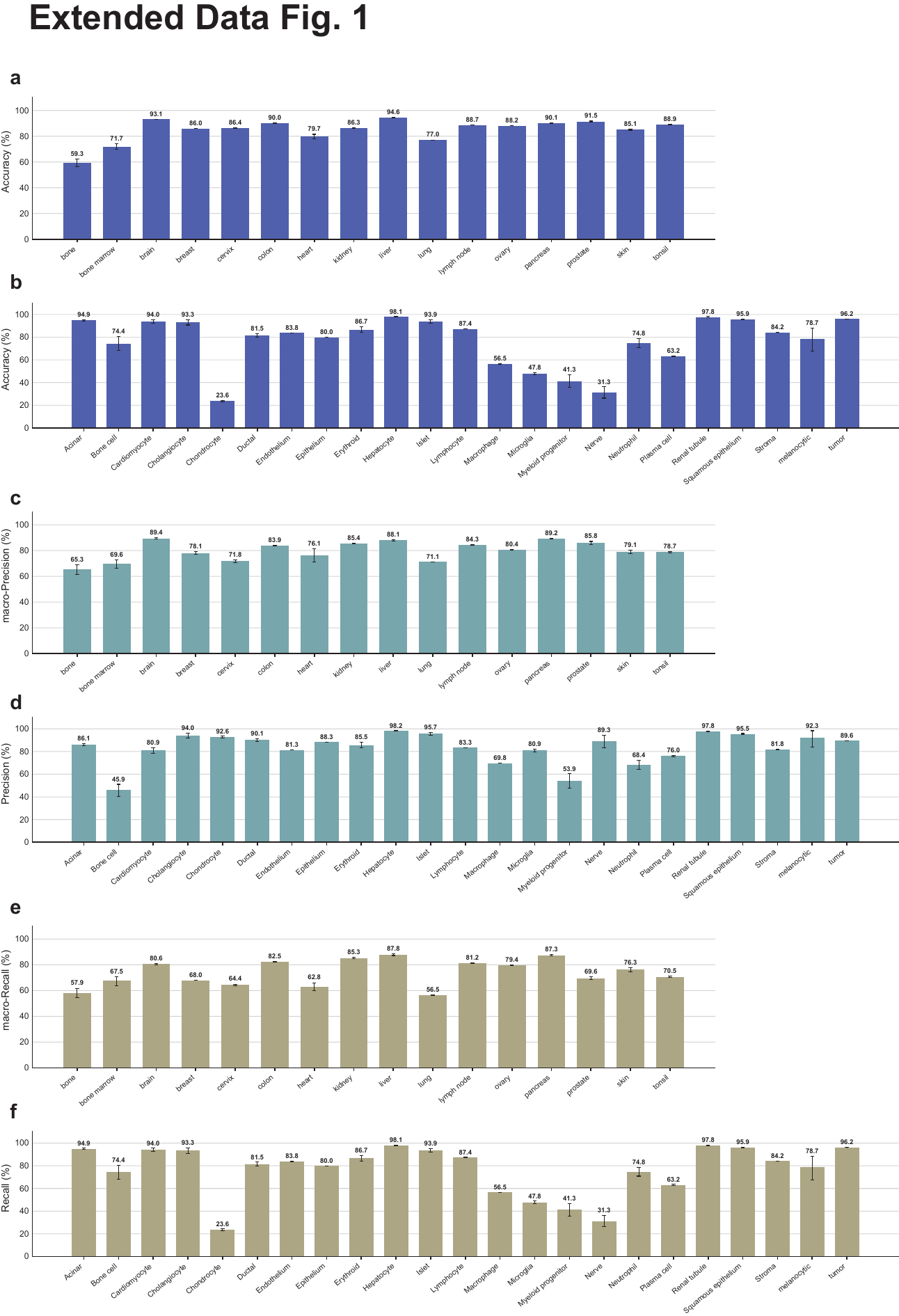}
\end{figure}


\begin{figure}[!t]
        \captionsetup{
          format=plain,
          justification=raggedright,
          singlelinecheck=false,
          margin=0pt,
          width=\dimexpr\textwidth\relax
        }
    \caption[\textbf{Additional in-distribution metrics on the held-out Xenium test split.}]%
    {\textbf{Additional in-distribution metrics on the held-out Xenium test split.}
    Complementary metrics to Figure~\ref{fig:figure2}a,b, all computed on the spatially held-out split ($2{,}944{,}356$ cells).
    \textbf{(a)} Accuracy per organ.
    \textbf{(b)} Accuracy per cell type, which for a single class equals its recall and is therefore identical to panel~\textbf{(f)}.
    \textbf{(c)} Macro-precision per organ.
    \textbf{(d)} Precision per cell type.
    \textbf{(e)} Macro-recall per organ.
    \textbf{(f)} Recall per cell type.
    Error bars represent $95$\% confidence intervals from bootstrap resampling over test cells.}
    \label{fig:extfigure1}
\end{figure}

\begin{figure}[!t]
    \centering
    \includegraphics[width=0.95\textwidth]{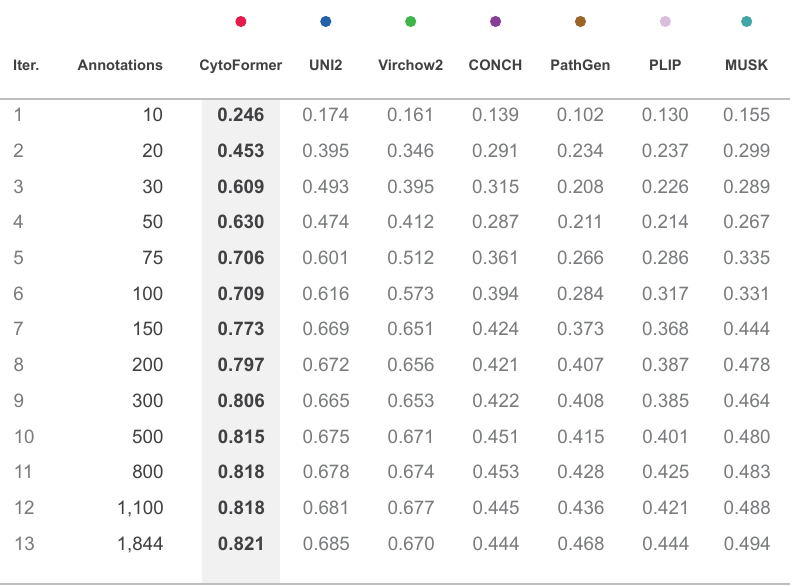}
        \captionsetup{
          format=plain,
          justification=raggedright,
          singlelinecheck=false,
          margin=0pt,
          width=\dimexpr\textwidth\relax
        }
    \caption{\textbf{Epithelium detection F1 across active-learning iterations.} Epithelium vs all others; 10,627 positive / 151,075 negative cells; same annotations, schedule and 5 seeds for every model. F1 for the Epithelium class, mean of 5 seeds.  Bold = best in row.}
    \label{fig:extfigure2}
\end{figure}

\section{Discussion}
\label{sec:discussion}

We have presented CytoFormer, a cell foundation model trained on \HandE{} cells whose identity comes from paired spatial transcriptomics. Reference labels were derived from Xenium by clustering, marker-gene annotation and human review, and were then transferred to the matched \HandE{} morphology. This gave $15.4$ million cells from $81$ sections and $16$ organs without any manual annotation of images. On spatially held-out tissue the model reached an accuracy of $0.846$ and a macro-F1 of $0.779$ across all $16$ organs, and when applied to a whole section its predictions reproduced the tissue architecture of the reference. The representation also transfers. Under linear fine-tuning it performed very well on four public benchmarks, including on organs and cell types that were not part of pretraining, and it continued to perform well when every model was fully fine-tuned.

Several limitations should be noted. First, our label space covers the cell types that are commonly reported in each organ, which are the ones used in routine practice, and does not go below that level. Finer subtypes are therefore not separated, for example T and B cells, which are grouped as Lymphocyte, or the states of macrophages, even though the transcriptome that produced the labels often carries that information. Second, this work uses Xenium alone as the source of cell types. In future work we will extend it to other paired modalities, such as spatial proteomics~\cite{shaban2025foundation} and Visium HD~\cite{oliveira2025visiumhd}. Third, we have used CytoFormer only to assign a type to each cell, but its output is a map of every cell in a slide, and that map is itself a measurement. Cell composition and the spatial arrangement of cells carry diagnostic and prognostic information, and in future work we will use the counts, densities and neighbourhoods that the model produces as features for clinical tasks such as diagnosis, prognosis and the prediction of treatment response~\cite{Li2025-kj}.

In summary, CytoFormer shows that paired spatial transcriptomics can replace manual annotation as the source of cell-level supervision on \HandE{}, and that a cell foundation model trained in this way transfers to organs and cell types it has never seen. The model turns any routine \HandE{} slide into a map of typed cells, and provides a cell-level representation that downstream analyses can build on.

\section{Methods}
\label{sec:methods}
\subsection{Dataset introduction}

\paragraph{Data sources.} We went through the human Xenium experiments available from two sources: the 10x Genomics public dataset portal, and HEST-1k~\cite{jaume2024hest}, which collects spatial transcriptomics datasets deposited elsewhere. We kept the experiments in which the same tissue section had also been imaged with \HandE{}, and removed samples that appear in both sources. After removing sections without an \HandE{} image or without a usable image-to-transcriptome registration, $81$ sections from $16$ organs entered training: bone, bone marrow, brain, breast, cervix, colon, heart, kidney, liver, lung, lymph node, ovary, pancreas, prostate, skin and tonsil. Of these, $54$ come from the 10x Genomics portal and $27$ from HEST-1k deposits. Gene panels range from targeted about 300-plex to the $5{,}000$-plex Prime panel, and sections include healthy, inflammatory and malignant tissue.

\paragraph{Reference cell typing.} Each section was processed independently in Scanpy~\cite{wolf2018scanpy}: counts were normalised per cell and $\log(1+x)$ transformed, highly variable genes selected, a PCA computed, and a $k$-nearest-neighbour graph clustered with Leiden~\cite{traag2019leiden} at a resolution of $0.8$. Sections sharing a gene panel within an organ were integrated before clustering; single sections were clustered on their own without scaling, so that absolute expression levels remain interpretable. For every cluster we computed ranked marker genes by Wilcoxon rank-sum test and annotated the cluster from the top $30$ markers together with the mean expression and per cent positivity of canonical lineage markers.

\paragraph{Quality control.} Two orthogonal filters were applied. At the cluster level, every cluster of every organ was reviewed by hand: whether the cell types recovered and their proportions match what the tissue should contain, what the top markers of the cluster are, and how its absolute expression compares with competing lineages. Each cluster was then kept, relabelled, or removed. We removed the clusters that were artefacts, doublets or of low quality, and flagged the clusters that are real but ambiguous, which are excluded from training. At the cell level, we computed a confidence for every cell, which combines two scores, a neighbour purity and a marker margin: the first asks whether a cell really belongs to the cluster it was assigned to, the second whether that cluster's markers are stronger in the cell than those of the most similar other cluster. Neighbour purity is the fraction of a cell's $k=50$ nearest neighbours in expression PCA space that belong to its own Leiden cluster. Marker margin is the mean expression of its own cluster's marker set minus that of the strongest competing cluster's marker set, and is then normalised to $[0,1]$. The confidence of a cell is the average of the two, and we kept only the cells with a confidence above $0.6$ for training. The intention is to drop the ambiguous cells that sit at the border between clusters to reduce the label noise.

\paragraph{Cell-type taxonomy.} For each organ we selected the cell types that are commonly reported in that tissue, and mapped the cluster-level annotations onto them, giving $23$ categories in total. Seven pan-tissue classes (tumour, Stroma, Epithelium, Lymphocyte, Endothelium, Macrophage, Plasma cell) cover about $97$\% of cells; the other $16$ are organ-specific (Hepatocyte, Cholangiocyte, Acinar, Ductal, Islet, Renal tubule, Microglia, Chondrocyte, Squamous epithelium, Cardiomyocyte, Erythroid, Neutrophil, Myeloid progenitor, Bone cell, Nerve and melanocytic). The subtypes found by clustering were then merged into these predefined categories, for example T and B cells into Lymphocyte.

\paragraph{\HandE{} registration and patch extraction.} Cell coordinates were mapped into \HandE{} pixel space using the registration supplied with each section, and we checked the mapping by overlaying nuclear boundaries from the DAPI channel on the \HandE{} image. For each cell a $56\,\mu$m window centred on the nucleus centroid was cropped from the finest pyramid level and resized to $224\times224$ pixels. Crops falling outside tissue were removed with a tissue mask that also excludes tears and folds, and out-of-focus crops were removed by thresholding the Laplacian variance of the resized patch at $10$, a cut-off we set by visually inspecting the discarded patches.

\paragraph{Training and testing split.} For breast and lung, where many comparable sections exist, whole sections were held out ($7$ of $23$ and $7$ of $28$). For the remaining $14$ organs, training and test cells were separated by location within the section rather than at random: the section was recursively bisected at the median of the cell distribution into $16$ KD-tree blocks~\cite{bentley1975kdtree}, and dispersed blocks totalling about $20$\% of cells were assigned to test. Because every cell is represented by a $56\,\mu$m crop, two cells whose centroids lie closer than $56\,\mu$m share image content, so a test cell next to a training cell would already have been partly seen during training. We therefore discarded a guard band at every block boundary: any cell whose nearest neighbour of the opposite split lay within $112\,\mu$m was removed from both splits. The threshold is twice the crop width, so every retained training cell and every retained test cell are at least one full crop apart and their fields of view cannot overlap. This removed $366{,}227$ cells, $2.4$\% of the dataset. The resulting split contains $12{,}111{,}769$ training, $2{,}944{,}356$ test and $366{,}227$ guard-band cells.

\subsection{Model architecture and training}
CytoFormer is a shared image encoder followed by a multi-task, per-organ classification head. The encoder is a ViT-giant transformer~\cite{dosovitskiy2021vit} that takes a $224\times224$ input with a patch size of $14$ and returns a $1536$-dimensional embedding. It is initialised from UNI2-h~\cite{Chen2024-sl} and fine-tuned end-to-end, and its pooled class token is layer-normalised to give one embedding per cell.

The head holds $16$ linear classifiers, one per organ, each mapping the embedding to the cell types of that organ alone. Head sizes range from three to nine classes, which gives $103$ outputs in total. At each forward pass the organ identifier sends every cell to the head of its organ.

We trained with AdamW~\cite{loshchilov2019adamw} (weight decay $10^{-2}$), using separate learning rates for the backbone ($2\times10^{-5}$) and the heads ($10^{-3}$), one epoch of linear warm-up followed by cosine decay, gradient clipping at $1.0$, and a batch size of $256$ in bfloat16, under distributed data-parallel training on four NVIDIA B200 GPUs. Dropout of $0.25$ is applied to the embedding before the heads. Augmentation consisted of random horizontal and vertical flips, random rotation up to $180^{\circ}$, and colour jitter with brightness, contrast and saturation $0.1$ and hue $0.02$. Training ran for up to $10$ epochs with early stopping on held-out macro-F1 (patience $5$), evaluated at the end of each epoch on a capped subsample of the test split for speed. We kept the checkpoint with the best macro-F1, which occurred at epoch $5$, and re-evaluated it on the full test split.

\subsection{Field-of-view ablation}
To isolate the effect of the crop window we sampled one fixed set of cells ($500{,}000$ for training and $100{,}000$ for testing, stratified by organ and cell type over $10$ organs and $17$ cell types) and cropped that same set at $28$, $56$, $112$ and $224\,\mu$m, each resized to $224$ pixels. Every other element of the recipe was held constant, so the crop window, and therefore the effective resolution, is the only variable. Models were compared by macro-F1 over the $17$ classes.

\subsection{Public benchmarks}
We benchmarked against six pathology foundation models used as feature extractors (UNI2-h, Virchow2, MUSK, CONCH, PathGen and PLIP) on four expert-annotated datasets: PanNuke~\cite{gamper2019pannuke} ($186{,}836$ cells, $19$ organs, $4$ classes), CoNSeP~\cite{graham2019hovernet} ($23{,}400$ cells, colon, $6$ classes), MoNuSAC~\cite{verma2021monusac} ($46{,}901$ cells, $4$ organs, $4$ classes) and PUMA~\cite{schuiveling2025puma} ($95{,}389$ cells, melanoma skin, $9$ classes). Categories that do not correspond to a definite cell identity were removed before the benchmark was built: \emph{Dead cells} in PanNuke, \emph{Ambiguous} in MoNuSAC, \emph{apoptosis} in PUMA and the unspecified \emph{other} class in CoNSeP. This leaves four, four, nine and six classes respectively. Each dataset's own split was used where defined (PanNuke folds $1$--$2$ for training and fold $3$ for testing; the CoNSeP and MoNuSAC official splits) and PUMA was split by region of interest (about $20$\% held out). Every cell was cropped identically at $56\,\mu$m and resized to $224$ pixels.

In the linear fine-tuning regime, features were extracted with the frozen backbone, standardised with training statistics, and a single linear layer was trained per (dataset, organ) with class-balanced cross-entropy (AdamW~\cite{loshchilov2019adamw}, learning rate $10^{-3}$, weight decay $10^{-4}$, $300$ full-batch steps). In the fine-tuning regime the same backbone and head were trained end-to-end for $10$ epochs with backbone learning rate $2\times10^{-5}$ and head learning rate $10^{-3}$, cosine schedule with warm-up, bfloat16, and early stopping on test macro-F1 with patience $3$.

\subsection{Statistical analysis}
Statistical significance between methods was assessed using a paired two-sided bootstrap test~\cite{efron1979bootstrap} over the test cells, with $1{,}000$ resamples and a single set of resampling indices shared by all methods. Significance levels are denoted $^{*}p<0.05$, $^{**}p<0.01$ and $^{***}p<0.001$. Confidence intervals ($95$\% CI) were computed by bootstrap resampling over test cells.

\subsection{Label-efficiency benchmark}
We benchmarked CytoFormer against six pathology foundation models used as feature extractors, UNI2-h, Virchow2, CONCH, PathGen, PLIP and MUSK, on interactive normal-epithelium detection in a colorectal cancer VisiumHD section.

We assigned each segmented nucleus a cell type from the paired VisiumHD transcriptome, aggregating the
$2\,\mu$m-bin expression that fell within its contour and scoring curated marker sets, and kept only confident calls that met total-UMI and score-margin thresholds. Because benign and malignant epithelium express the same markers, we further resolved the benign versus malignant axis from crypt structure. Epithelial cells lying within normal crypts were labelled normal Epithelium, epithelial cells elsewhere were labelled Tumor, and all non-epithelial cells were labelled Negative control. This yielded $161{,}702$ confidently typed cells, of which $10{,}627$ were normal Epithelium and $151{,}075$ were other cells.

For every cell we cropped a $56\,\mu$m H\&E patch centred on its nucleus, resized it to $224\times224$
pixels and encoded it with each foundation model to obtain a per-cell embedding. On top of each embedding
we trained a multinomial logistic-regression head with balanced class weights and $C=1$ on the user
annotations and predicted the three classes for every cell. All models shared the same annotations, the
same cells and the same head, so the embedding was the only variable.

We scored normal-epithelium detection as the F1 of the Epithelium class, one versus all other cells, over all $161{,}702$ confidently typed cells. Because the goal is to measure how accurately a small set of annotations propagates across the slide under active learning rather than generalization to unseen data, we evaluated on the whole slide including the annotated cells, which form a small fraction of the evaluation set and are identical across models. To trace label efficiency we retrained on nested random subsets of the annotations of increasing size, from $10$ up to $1{,}844$, and averaged over five fixed random seeds ($0$ to $4$); at each annotation budget the same seed selected the same subset for every model, so the comparison is matched and reproducible.

\section*{Author Contributions}

J.Y., S.L. developed the methodology. J.Y., S.L., A.Y. performed the experiments. J.Y., S.L., A.Y., Z.H. wrote the manuscript. Z.H. conceived and supervised the study. All authors discussed the results and approved the final manuscript.

\section*{Acknowledgements}
This project was supported by startup funding from the Perelman School of Medicine, University of Pennsylvania (Z.H.) and the Abramson Cancer Center Pilot Award (Z.H.).



\section*{Code Availability}

Code is available at \url{https://github.com/zhihuanglab/CytoFormer}. Model checkpoint is available at \url{https://huggingface.co/zhihuanglab/CytoFormer}.

\section*{Conflict of Interests}
None declared.

\clearpage
\printbibliography
\end{document}

%% file: macros.tex
\usepackage[dvipsnames,svgnames, table,xcdraw]{xcolor}
\usepackage{relsize}
\usepackage{tcolorbox}
\usepackage{listings}
\usepackage{subcaption}
\usepackage{algorithm}
\usepackage{algorithmic}
\usepackage{comment}
\usepackage{graphicx}
\usepackage{booktabs}
\usepackage{multirow}
\usepackage{todonotes}
\usepackage{enumitem}
\usepackage{url}
\usepackage{mathpazo}
\usepackage{amsmath,amssymb,amsthm,amsfonts,bm}
\usepackage[toc,page,header]{appendix}
\usepackage{fancyhdr}
\usepackage[T1]{fontenc}

\usepackage{auth_detailed}

\counterwithout{figure}{section}
\counterwithout{table}{section}
\definecolor{darkgoldenrod}{rgb}{0.72, 0.53, 0.04}
\definecolor{backgroundcolor}{RGB}{250, 250, 252}   
\definecolor{keywordcolor}{RGB}{30, 0, 178}       
\definecolor{stringcolor}{RGB}{204, 0, 102}        
\definecolor{numbercolor}{RGB}{0, 128, 128}        
\definecolor{emphcolor}{RGB}{30, 0, 178}            
\definecolor{commentcolor}{RGB}{0, 128, 0}       
\definecolor{basiccodecolor}{RGB}{61, 61, 61}       

\lstdefinestyle{customstyle}{
    backgroundcolor=\color{backgroundcolor},   
    commentstyle=\color{commentcolor},
    keywordstyle=\color{keywordcolor},
    numberstyle=\color{numbercolor},
    stringstyle=\color{stringcolor},
    basicstyle=\color{basiccodecolor}\ttfamily\footnotesize,
    breakatwhitespace=false,         
    breaklines=true,                 
    captionpos=b,                    
    keepspaces=true,                 
    numbers=left,     
    basicstyle=\color{basiccodecolor}\ttfamily\footnotesize,
    numbersep=5pt,             
    xleftmargin=2em,
    xrightmargin=2em,
    showspaces=false,                
    showstringspaces=false,
    showtabs=false,                  
    tabsize=1,
    frame=single,
    framesep=5pt,
    framexleftmargin=1.5em,
    framexrightmargin=1.5em,
    framextopmargin=1pt,
    framexbottommargin=1pt,
    aboveskip=10pt,
    belowskip=10pt,
    breaklines=true,
    breakautoindent=true,
    emph={},
    emphstyle={\color{emphcolor}},
    extendedchars=true,
}

\input{editor}


%% file: editor.tex
\usepackage[normalem]{ulem}
\usepackage{xcolor}

\newcommand{\scrcmd}{\textsuperscript}
\newcommand{\scr}[1]{\scrcmd{#1}}

%% file: bib.bib
@ARTICLE{Huang2025-bv,
  title     = "{STPath}: a generative foundation model for integrating spatial
               transcriptomics and whole-slide images",
  author    = "Huang, Tinglin and Liu, Tianyu and Babadi, Mehrtash and Ying, Rex
               and Jin, Wengong",
  journal   = "NPJ Digit. Med.",
  publisher = "Springer Science and Business Media LLC",
  volume    =  8,
  number    =  1,
  pages     =  659,
  month     =  nov,
  year      =  2025,
  language  = "en"
}

@ARTICLE{Marco-Salas2025-xz,
  title     = "Optimizing Xenium In Situ data utility by quality assessment and
               best-practice analysis workflows",
  author    = "Marco Salas, Sergio and Kuemmerle, Louis B and Mattsson-Langseth,
               Christoffer and Tismeyer, Sebastian and Avenel, Christophe and
               Hu, Taobo and Rehman, Habib and Grillo, Marco and Czarnewski,
               Paulo and Helgadottir, Saga and Tiklova, Katarina and Andersson,
               Axel and Rafati, Nima and Chatzinikolaou, Maria and Theis, Fabian
               J and Luecken, Malte D and Wählby, Carolina and Ishaque, Naveed
               and Nilsson, Mats",
  journal   = "Nat. Methods",
  publisher = "Nature Publishing Group",
  volume    =  22,
  number    =  4,
  pages     = "813--823",
  month     =  apr,
  year      =  2025,
  language  = "en"
}

@ARTICLE{Xu2024-do,
  title     = "A whole-slide foundation model for digital pathology from
               real-world data",
  author    = "Xu, Hanwen and Usuyama, Naoto and Bagga, Jaspreet and Zhang,
               Sheng and Rao, Rajesh and Naumann, Tristan and Wong, Cliff and
               Gero, Zelalem and González, Javier and Gu, Yu and Xu, Yanbo and
               Wei, Mu and Wang, Wenhui and Ma, Shuming and Wei, Furu and Yang,
               Jianwei and Li, Chunyuan and Gao, Jianfeng and Rosemon, Jaylen
               and Bower, Tucker and Lee, Soohee and Weerasinghe, Roshanthi and
               Wright, Bill J and Robicsek, Ari and Piening, Brian and Bifulco,
               Carlo and Wang, Sheng and Poon, Hoifung",
  journal   = "Nature",
  publisher = "Springer Science and Business Media LLC",
  volume    =  630,
  number    =  8015,
  pages     = "181--188",
  month     =  jun,
  year      =  2024,
  language  = "en"
}

@ARTICLE{Wang2024-hs,
  title     = "A pathology foundation model for cancer diagnosis and prognosis
               prediction",
  author    = "Wang, Xiyue and Zhao, Junhan and Marostica, Eliana and Yuan, Wei
               and Jin, Jietian and Zhang, Jiayu and Li, Ruijiang and Tang,
               Hongping and Wang, Kanran and Li, Yu and Wang, Fang and Peng,
               Yulong and Zhu, Junyou and Zhang, Jing and Jackson, Christopher R
               and Zhang, Jun and Dillon, Deborah and Lin, Nancy U and Sholl,
               Lynette and Denize, Thomas and Meredith, David and Ligon, Keith L
               and Signoretti, Sabina and Ogino, Shuji and Golden, Jeffrey A and
               Nasrallah, MacLean P and Han, Xiao and Yang, Sen and Yu,
               Kun-Hsing",
  journal   = "Nature",
  publisher = "Springer Science and Business Media LLC",
  volume    =  634,
  number    =  8035,
  pages     = "970--978",
  month     =  oct,
  year      =  2024,
  language  = "en"
}

@ARTICLE{Chen2024-sl,
  title     = "Towards a general-purpose foundation model for computational
               pathology",
  author    = "Chen, Richard J and Ding, Tong and Lu, Ming Y and Williamson,
               Drew F K and Jaume, Guillaume and Song, Andrew H and Chen, Bowen
               and Zhang, Andrew and Shao, Daniel and Shaban, Muhammad and
               Williams, Mane and Oldenburg, Lukas and Weishaupt, Luca L and
               Wang, Judy J and Vaidya, Anurag and Le, Long Phi and Gerber,
               Georg and Sahai, Sharifa and Williams, Walt and Mahmood, Faisal",
  journal   = "Nat. Med.",
  publisher = "Nature Publishing Group",
  volume    =  30,
  number    =  3,
  pages     = "850--862",
  month     =  mar,
  year      =  2024,
  language  = "en"
}

@ARTICLE{Xiang2025-vd,
  title     = "A vision--language foundation model for precision oncology",
  author    = "Xiang, Jinxi and Wang, Xiyue and Zhang, Xiaoming and Xi, Yinghua
               and Eweje, Feyisope and Chen, Yijiang and Li, Yuchen and
               Bergstrom, Colin and Gopaulchan, Matthew and Kim, Ted and Yu,
               Kun-Hsing and Willens, Sierra and Olguin, Francesca Maria and
               Nirschl, Jeffrey J and Neal, Joel and Diehn, Maximilian and Yang,
               Sen and Li, Ruijiang",
  journal   = "Nature",
  publisher = "Springer Science and Business Media LLC",
  volume    =  638,
  number    =  8051,
  pages     = "769--778",
  month     =  feb,
  year      =  2025,
  language  = "en"
}

@ARTICLE{Chen2025-ee,
  title     = "A visual--omics foundation model to bridge histopathology with
               spatial transcriptomics",
  author    = "Chen, Weiqing and Zhang, Pengzhi and Tran, Tu N and Xiao, Yiwei
               and Li, Shengyu and Shah, Vrutant V and Cheng, Hao and Brannan,
               Kristopher W and Youker, Keith and Lai, Li and Fang, Longhou and
               Yang, Yu and Le, Nhat-Tu and Abe, Jun-ichi and Chen, Shu-Hsia and
               Ma, Qin and Chen, Ken and Song, Qianqian and Cooke, John P and
               Wang, Guangyu",
  journal   = "Nat. Methods",
  publisher = "Springer Science and Business Media LLC",
  volume    =  22,
  number    =  7,
  pages     = "1568--1582",
  month     =  jul,
  year      =  2025,
  language  = "en"
}

@ARTICLE{Huang2023-di,
  title     = "A visual--language foundation model for pathology image analysis
               using medical Twitter",
  author    = "Huang, Zhi and Bianchi, Federico and Yuksekgonul, Mert and
               Montine, Thomas J and Zou, James",
  journal   = "Nat. Med.",
  publisher = "Springer Science and Business Media LLC",
  volume    =  29,
  number    =  9,
  pages     = "2307--2316",
  month     =  sep,
  year      =  2023,
  language  = "en"
}

@ARTICLE{Li2025-kj,
  title         = "A co-evolving agentic {AI} system for medical imaging
                   analysis",
  author        = "Li, Songhao and Xu, Jonathan and Bao, Tiancheng and Liu,
                   Yuxuan and Liu, Yuchen and Liu, Yihang and Wang, Lilin and
                   Lei, Wenhui and Wang, Sheng and Xu, Yinuo and Cui, Yan and
                   Yao, Jialu and Koga, Shunsuke and Huang, Zhi",
  journal       = "arXiv [cs.CV]",
  month         =  sep,
  year          =  2025,
  archivePrefix = "arXiv",
  primaryClass  = "cs.CV"
}

@ARTICLE{Lu2024-dz,
  title     = "A visual-language foundation model for computational pathology",
  author    = "Lu, Ming Y and Chen, Bowen and Williamson, Drew F K and Chen,
               Richard J and Liang, Ivy and Ding, Tong and Jaume, Guillaume and
               Odintsov, Igor and Le, Long Phi and Gerber, Georg and Parwani,
               Anil V and Zhang, Andrew and Mahmood, Faisal",
  journal   = "Nat. Med.",
  publisher = "Nature Publishing Group",
  volume    =  30,
  number    =  3,
  pages     = "863--874",
  month     =  mar,
  year      =  2024,
  language  = "en"
}

@article{shaban2025foundation,
  title={A foundation model for spatial proteomics},
  author={Shaban, Muhammad and Chang, Yuzhou and Qiu, Huaying and Yeo, Yao Yu and Song, Andrew H and Jaume, Guillaume and Wang, Yuchen and Weishaupt, Luca L and Ding, Tong and Vaidya, Anurag and others},
  journal={arXiv preprint arXiv:2506.03373},
  year={2025}
}

@inproceedings{jaume2024hest,
  title     = {{HEST-1k}: A Dataset for Spatial Transcriptomics and Histology Image Analysis},
  author    = {Jaume, Guillaume and Doucet, Paul and Song, Andrew H. and Lu, Ming Y. and Almagro-P{\'e}rez, Cristina and Wagner, Sophia J. and Vaidya, Anurag J. and Chen, Richard J. and Williamson, Drew F. K. and Kim, Ahrong and Mahmood, Faisal},
  booktitle = {Advances in Neural Information Processing Systems (NeurIPS), Datasets and Benchmarks Track},
  year      = {2024},
  url       = {https://arxiv.org/abs/2406.16192}
}

@article{janesick2023xenium,
  title   = {High resolution mapping of the tumor microenvironment using integrated single-cell, spatial and in situ analysis},
  author  = {Janesick, Amanda and Shelansky, Robert and Gottscho, Andrew D. and Wagner, Florian and Williams, Stephen R. and Rouault, Morgane and Beliakoff, Ghezal and Morrison, Carolyn A. and Oliveira, Michelli F. and Sicherman, Jordan T. and Kohlway, Andrew and Abousoud, Jawad and Drennon, Tingsheng Yu and Mohabbat, Seayar H. and {10x Development Teams} and Taylor, Sarah E. B.},
  journal = {Nature Communications},
  volume  = {14},
  pages   = {8353},
  year    = {2023}
}

@article{zimmermann2024virchow2,
  title   = {{Virchow2}: Scaling Self-Supervised Mixed Magnification Models in Pathology},
  author  = {Zimmermann, Eric and Vorontsov, Eugene and Viret, Julian and Casson, Adam and Zelechowski, Michal and Shaikovski, George and Tenenholtz, Neil and Hall, James and Klimstra, David and Yousfi, Razik and Fuchs, Thomas and Fusi, Nicolo and Liu, Siqi and Severson, Kristen},
  journal = {arXiv preprint arXiv:2408.00738},
  year    = {2024}
}

@inproceedings{sun2025pathgen,
  title     = {{PathGen-1.6M}: 1.6 Million Pathology Image-text Pairs Generation through Multi-agent Collaboration},
  author    = {Sun, Yuxuan and Zhang, Yunlong and Si, Yixuan and Zhu, Chenglu and Zhang, Kai and Shui, Zhongyi and Li, Jingxiong and Gong, Xuan and Lyu, Xinheng and Lin, Tao and Yang, Lin},
  booktitle = {International Conference on Learning Representations (ICLR)},
  year      = {2025},
  url       = {https://openreview.net/forum?id=rFpZnn11gj}
}

@inproceedings{gamper2019pannuke,
  title     = {{PanNuke}: An Open Pan-Cancer Histology Dataset for Nuclei Instance Segmentation and Classification},
  author    = {Gamper, Jevgenij and Alemi Koohbanani, Navid and Benet, Ksenija and Khuram, Ali and Rajpoot, Nasir},
  booktitle = {European Congress on Digital Pathology (ECDP)},
  pages     = {11--19},
  year      = {2019}
}

@article{graham2019hovernet,
  title   = {{Hover-Net}: Simultaneous segmentation and classification of nuclei in multi-tissue histology images},
  author  = {Graham, Simon and Vu, Quoc Dang and Raza, Shan E Ahmed and Azam, Ayesha and Tsang, Yee Wah and Kwak, Jin Tae and Rajpoot, Nasir},
  journal = {Medical Image Analysis},
  volume  = {58},
  pages   = {101563},
  year    = {2019}
}

@article{verma2021monusac,
  title   = {{MoNuSAC2020}: A Multi-Organ Nuclei Segmentation and Classification Challenge},
  author  = {Verma, Ruchika and Kumar, Neeraj and Patil, Abhijeet and Kurian, Nikhil Cherian and Rane, Swapnil and Graham, Simon and Vu, Quoc Dang and Zwager, Mieke and Raza, Shan E Ahmed and Rajpoot, Nasir and others},
  journal = {IEEE Transactions on Medical Imaging},
  volume  = {40},
  number  = {12},
  pages   = {3413--3423},
  year    = {2021}
}

@article{schuiveling2025puma,
  title   = {A novel dataset for nuclei and tissue segmentation in melanoma with baseline nuclei segmentation and tissue segmentation benchmarks},
  author  = {Schuiveling, Mark and Liu, Hong and Eek, Daniel and Breimer, Gerben E. and Suijkerbuijk, Karijn P. M. and Blokx, Willeke A. M. and Veta, Mitko},
  journal = {GigaScience},
  volume  = {14},
  pages   = {giaf011},
  year    = {2025},
  doi     = {10.1093/gigascience/giaf011}
}

@article{traag2019leiden,
  title   = {From {Louvain} to {Leiden}: guaranteeing well-connected communities},
  author  = {Traag, Vincent A. and Waltman, Ludo and van Eck, Nees Jan},
  journal = {Scientific Reports},
  volume  = {9},
  pages   = {5233},
  year    = {2019}
}

@article{wolf2018scanpy,
  title   = {{SCANPY}: large-scale single-cell gene expression data analysis},
  author  = {Wolf, F. Alexander and Angerer, Philipp and Theis, Fabian J.},
  journal = {Genome Biology},
  volume  = {19},
  pages   = {15},
  year    = {2018}
}

@inproceedings{loshchilov2019adamw,
  title     = {Decoupled Weight Decay Regularization},
  author    = {Loshchilov, Ilya and Hutter, Frank},
  booktitle = {International Conference on Learning Representations (ICLR)},
  year      = {2019}
}

@article{bentley1975kdtree,
  title   = {Multidimensional binary search trees used for associative searching},
  author  = {Bentley, Jon Louis},
  journal = {Communications of the ACM},
  volume  = {18},
  number  = {9},
  pages   = {509--517},
  year    = {1975}
}

@inproceedings{dosovitskiy2021vit,
  title     = {An Image is Worth 16x16 Words: Transformers for Image Recognition at Scale},
  author    = {Dosovitskiy, Alexey and Beyer, Lucas and Kolesnikov, Alexander and Weissenborn, Dirk and Zhai, Xiaohua and Unterthiner, Thomas and Dehghani, Mostafa and Minderer, Matthias and Heigold, Georg and Gelly, Sylvain and Uszkoreit, Jakob and Houlsby, Neil},
  booktitle = {International Conference on Learning Representations (ICLR)},
  year      = {2021}
}

@article{oliveira2025visiumhd,
  title   = {High-definition spatial transcriptomic profiling of immune cell populations in colorectal cancer},
  author  = {Oliveira, Michelli Faria de and Romero, Juan Pablo and Chung, Meii and Williams, Stephen R. and Gottscho, Andrew D. and Gupta, Anushka and Pilipauskas, Susan E. and Mohabbat, Seayar and Raman, Nandhini and Sukovich, David J. and Patterson, David M. and {Visium HD Development Team} and Taylor, Sarah E. B.},
  journal = {Nature Genetics},
  volume  = {57},
  pages   = {1512--1523},
  year    = {2025},
  doi     = {10.1038/s41588-025-02193-3}
}

@article{efron1979bootstrap,
  title   = {Bootstrap Methods: Another Look at the Jackknife},
  author  = {Efron, Bradley},
  journal = {The Annals of Statistics},
  volume  = {7},
  number  = {1},
  pages   = {1--26},
  year    = {1979}
}

@article{yao2026melandx,
  title   = {{Melan-Dx}: a knowledge-enhanced vision-language framework improves differential diagnosis of melanocytic neoplasm pathology},
  author  = {Yao, Jialu and Li, Songhao and Liang, Peixian and Xu, Xiaowei and Elder, David and Huang, Zhi},
  journal = {npj Digital Medicine},
  volume  = {9},
  number  = {1},
  pages   = {171},
  year    = {2026},
  doi     = {10.1038/s41746-026-02357-3}
}
